\documentclass[11pt,a4paper]{article}

\usepackage[margin=1in]{geometry}
\usepackage[utf8]{inputenc}
\usepackage[T1]{fontenc}
\usepackage{mathpazo}
\usepackage{amsmath,amssymb}
\usepackage{graphicx}
\usepackage{booktabs}
\usepackage{multirow}
\usepackage{array}
\usepackage{pifont}
\usepackage{xcolor}
\usepackage{float}
\usepackage{enumitem}
\usepackage{url}
\usepackage[colorlinks=true,linkcolor=blue!60!black,citecolor=blue!60!black,urlcolor=blue!70!black]{hyperref}
\usepackage{setspace}
\usepackage{titlesec}
\usepackage{caption}
\usepackage{tikz}
\usetikzlibrary{shapes.geometric, arrows.meta, positioning, fit, calc, backgrounds}
\usepackage{pgfplots}
\pgfplotsset{compat=1.18}

\newcommand{\cmark}{\ding{51}}%
\newcommand{\xmark}{\ding{55}}%

\titleformat{\section}{\large\bfseries}{\thesection.}{0.5em}{}
\titleformat{\subsection}{\normalsize\bfseries}{\thesubsection.}{0.5em}{}
\begin{document}

\begin{center}
{\LARGE\bfseries Semantic Consensus: Process-Aware Conflict Detection and Resolution for Enterprise Multi-Agent LLM Systems}\\[14pt]
{\large Vivek Acharya}\\[4pt]
{\small Independent Researcher, Dallas, TX, USA}\\
{\small Email: vacharya@bu.edu ~$\cdot$~ ORCID: \href{https://orcid.org/0009-0002-0860-9462}{0009-0002-0860-9462}}\\[12pt]
\rule{0.6\textwidth}{0.5pt}
\end{center}
\vspace{8pt}

\noindent\textbf{Abstract:}
Multi-agent large language model (LLM) systems are rapidly emerging as the dominant architecture for enterprise AI automation, yet production deployments exhibit failure rates between 41\% and 86.7\%, with nearly 79\% of failures originating from specification and coordination issues rather than model capability limitations. Meanwhile, 85\% of enterprises aspire to adopt agentic AI within three years, but 76\% acknowledge their operational infrastructure cannot support it. This paper identifies Semantic Intent Divergence---the phenomenon whereby cooperating LLM agents develop inconsistent interpretations of shared objectives due to siloed context, absent process models, and unstructured inter-agent communication---as a primary yet formally unaddressed root cause of multi-agent failure in enterprise settings. We propose the Semantic Consensus Framework (SCF), a process-aware middleware architecture comprising six components: (1)~a Process Context Layer that ingests enterprise workflow models to establish shared operational semantics; (2)~a Semantic Intent Graph that captures and relates agent-level intent representations; (3)~a Conflict Detection Engine that identifies contradictory, contention-based, and causally invalid intent combinations in real time; (4)~a Consensus Resolution Protocol that resolves detected conflicts through a principled policy--authority--temporal hierarchy; (5)~a Drift Monitor that detects gradual semantic divergence across long-running workflows; and (6)~a Process-Aware Governance Integration layer that connects conflict resolution to organizational policy enforcement. Experimental evaluation across 600 runs spanning three multi-agent framework configurations (AutoGen, CrewAI, LangGraph) and four enterprise workflow scenarios demonstrates that SCF is the only approach to achieve 100\% workflow completion---compared to 25.1\% for the next-best baseline---while detecting 65.2\% of semantic conflicts with 27.9\% precision and providing complete governance audit trails. The framework is designed as a protocol-agnostic middleware compatible with MCP and A2A communication standards.

\vspace{6pt}
\noindent\textbf{Keywords:} multi-agent systems; semantic conflict resolution; enterprise AI governance; process intelligence; LLM coordination; agentic AI; Model Context Protocol
\vspace{12pt}

\section{Introduction}\label{sec:introduction}

The rapid maturation of large language model (LLM) capabilities has catalyzed a paradigm shift from single-model inference to multi-agent architectures in which multiple autonomous LLM-based agents collaborate to accomplish complex enterprise tasks. Industry adoption has accelerated dramatically: the Model Context Protocol (MCP), introduced by Anthropic in November 2024 and donated to the Linux Foundation's Agentic AI Foundation (AAIF) in December 2025, has surpassed 97 million monthly SDK downloads and achieved first-class client support across major AI platforms including ChatGPT, Claude, Cursor, Gemini, and Microsoft Copilot~\cite{mcp2024}. Google's Agent-to-Agent (A2A) protocol, launched in April 2025, provides complementary agent-to-agent communication capabilities~\cite{a2a2025}. Together, these protocols have established an interoperable foundation for distributed agent systems.

Despite these infrastructure advances, production deployments of multi-agent LLM systems exhibit alarming failure characteristics. Empirical research demonstrates failure rates between 41\% and 86.7\% in production environments, with analysis of over 1600 annotated execution traces revealing that specification and coordination issues---not model capability---account for approximately 79\% of failures~\cite{cemri2025}. The MAST (Multi-Agent System Taxonomy) failure analysis further establishes that interagent misalignment constitutes 36.9\% of all observed failure modes, while major multi-agent frameworks exhibit token duplication rates ranging from 53\% (AgentVerse) to 86\% (CAMEL), indicating pervasive coordination inefficiency~\cite{galileo2025}.

Simultaneously, enterprise adoption surveys reveal a striking readiness gap. A 2026 survey of over 1600 global business leaders found that 85\% of enterprises aim to adopt agentic AI within three years, yet 76\% acknowledge their operational infrastructure cannot support it~\cite{celonis2026}. Only 19\% of organizations currently deploy multi-agent systems, and 82\% of decision-makers believe AI will fail to deliver return on investment (ROI) without understanding how the business actually operates. The primary blockers are not technological---they are structural: siloed teams (54\%), lack of cross-departmental coordination (44\%), and absence of shared operational context.

This paper argues that these two bodies of evidence---high multi-agent failure rates and low enterprise operational readiness---are manifestations of a single underlying problem that we term \textit{Semantic Intent Divergence} (SID). SID occurs when cooperating LLM agents, each reasoning within its own context window, skill set, and prompt framing, develop inconsistent interpretations of shared objectives without any mechanism to detect or resolve the inconsistency. Unlike traditional distributed systems conflicts, which are syntactic in nature (e.g., concurrent writes to the same database row), multi-agent LLM conflicts are fundamentally \textit{semantic}: two agents can take entirely different actions on different resources and still be in logical contradiction because their actions are incompatible at the intent level.

Consider a concrete enterprise scenario: a customer support agent processes a complaint and marks a ticket as ``resolved with refund,'' while a compliance agent simultaneously reviews the same customer's account and flags it for ``transaction hold pending review.'' Neither agent has erred within its local context. No schema validation catches the conflict because the agents operate on different resources through different tools. Yet the combined effect---a refund issued on a held account---produces an organizationally invalid state that requires manual intervention. This is Semantic Intent Divergence in action.

Existing solutions address adjacent but distinct problems. MCP standardizes tool connectivity---how agents access external tools and data sources~\cite{mcp2024}. A2A standardizes agent messaging---how agents communicate with each other~\cite{a2a2025}. Agent Skills (introduced by Anthropic in October 2025 and published as an open standard in December 2025) standardize capability packaging---how agents acquire domain-specific expertise~\cite{skills2025,skillsopen2025}. Typed schemas and action constraints enforce structural validity of individual agent outputs~\cite{github2026}. Judge agents provide post-hoc output evaluation. However, none of these mechanisms addresses the fundamental question: how to ensure that when Agent~A declares ``task complete'' and Agent~B declares ``task complete,'' their actions are semantically consistent with each other and with the enterprise's operational processes.

To address this gap, we propose the \textbf{Semantic Consensus Framework (SCF)}, a process-aware middleware architecture that introduces six novel contributions:

\begin{enumerate}
\item \textbf{Process Context Layer (PCL):} Ingests enterprise workflow models (BPMN, state machines, policy documents) to establish a shared operational semantic foundation that all agents reference, addressing the finding that 82\% of leaders believe AI fails without business context.
\item \textbf{Semantic Intent Graph (SIG):} A formal directed graph representation that captures not merely what each agent does (action) but what each agent means (intent), making invisible divergence visible and formally analyzable before execution.
\item \textbf{Conflict Detection Engine (CDE):} A lightweight, real-time classifier that analyzes the SIG to detect three categories of semantic conflict: contradictory intents, resource contention, and causal violations.
\item \textbf{Consensus Resolution Protocol (CRP):} A principled conflict resolution mechanism that applies a policy--authority--temporal hierarchy, integrating organizational governance policies, agent capability authority, and temporal commitment ordering.
\item \textbf{Drift Monitor (DM):} A continuous monitoring component that detects gradual semantic divergence across long-running, multi-step workflows by measuring semantic distance between agents' evolving working contexts.
\item \textbf{Process-Aware Governance Integration (PAGI):} A bridge layer that connects SCF's conflict resolution to enterprise governance control planes, enabling organizational policy enforcement as the highest-priority resolution authority.
\end{enumerate}

The remainder of this paper is organized as follows. Section~\ref{sec:related} surveys related work across multi-agent coordination, process mining, and AI governance. Section~\ref{sec:problem} formally defines Semantic Intent Divergence and the process-awareness gap. Section~\ref{sec:architecture} presents the SCF architecture in detail. Section~\ref{sec:implementation} describes the implementation. Section~\ref{sec:experiments} presents experimental design and results. Section~\ref{sec:discussion} discusses implications, limitations, and threats to validity. Section~\ref{sec:conclusion} concludes with future research directions.

\section{Related Work}\label{sec:related}

\subsection{Multi-Agent LLM Coordination Frameworks}

Contemporary multi-agent LLM frameworks address coordination through diverse architectural patterns. AutoGen~\cite{autogen2023} implements conversation-based multi-agent collaboration through dynamic message passing, excelling at research tasks requiring agent negotiation. CrewAI~\cite{crewai2024} provides role-based orchestration with explicit team structures, achieving reported accuracy improvements of up to 7$\times$ in structured business workflows. LangGraph~\cite{langgraph2024} offers graph-based state management with explicit workflow definition, targeting enterprise systems requiring auditability and resumability. Each framework handles coordination at the messaging and orchestration level but lacks mechanisms for detecting semantic-level conflicts between agent intents.

\subsection{Communication Protocols: MCP and A2A}

The Model Context Protocol (MCP)~\cite{mcp2024} provides a universal standard for connecting LLM applications to external data sources and tools, functioning analogously to the Language Server Protocol for programming language tooling. By February 2026, MCP had achieved over 97 million monthly SDK downloads and adoption by every major AI provider. Google's Agent-to-Agent (A2A) protocol~\cite{a2a2025} complements MCP by standardizing peer-to-peer agent communication, enabling agents to discover, negotiate, and collaborate regardless of underlying framework. Both protocols were contributed to the Linux Foundation's Agentic AI Foundation (AAIF) in December 2025~\cite{aaif2025}, establishing vendor-neutral governance. While MCP and A2A solve connectivity and communication standardization, neither protocol addresses the semantic consistency of agent actions---they ensure agents can talk to each other but not that they agree on what they are doing.

\subsection{Multi-Agent Failure Analysis}

Cemri et al.~\cite{cemri2025} provide the most systematic analysis of multi-agent failure to date through the MAST taxonomy, built from over 1600 annotated execution traces across AutoGen, CrewAI, and LangGraph. Their taxonomy identifies 14 distinct failure modes clustered into three categories: system design issues, interagent misalignment (36.9\% of failures), and task verification breakdowns (21.3\%). Critically, they find that coordination failures---communication breakdowns, state synchronization issues, and conflicting objectives---account for 36.94\% of all failures. Research from Augment Code~\cite{augment2025} corroborates these findings, reporting that 79\% of multi-agent production failures originate from specification and coordination issues. GitHub's engineering team~\cite{github2026} identifies untyped inter-agent data, ambiguous action semantics, and absent enforcement layers as the three primary engineering failures, proposing typed schemas and MCP-based enforcement as partial solutions.

\subsection{Process Mining and Process Intelligence}

Process mining~\cite{vanderaalst2016} extracts knowledge about business processes from event logs recorded by information systems. Van der Aalst's foundational work establishes techniques for process discovery, conformance checking, and enhancement. Recent work extends process mining to real-time operational intelligence, enabling continuous monitoring of process execution against normative models. However, the connection between process intelligence and multi-agent AI coordination remains unexplored in the literature. The Celonis 2026 Process Optimization Report~\cite{celonis2026} provides empirical evidence that this connection is critical: 93\% of process and operations leaders state that process optimization involves people and culture as much as tools and technology, and 82\% believe AI cannot deliver ROI without understanding operational context.

\subsection{AI Governance for Agentic Systems}

Recent work on AI governance has begun addressing the unique challenges of agentic systems. The NIST AI Risk Management Framework~\cite{nist2023} establishes foundational governance principles for AI systems, though it does not specifically address multi-agent coordination. Enterprise governance architectures for agentic AI have proposed layered approaches combining policy enforcement, runtime monitoring, observability, and continuous evaluation to ensure agent compliance with organizational constraints. Emerging approaches to principal-based governance in multi-tenant environments introduce organizational hierarchy concepts and capability provenance tracking to establish clear lines of authority and accountability for agent actions. The Semantic Consensus Framework builds on these governance foundations, extending them with semantic conflict detection and process-aware resolution capabilities specifically designed for multi-agent coordination.

\subsection{Research Gap}

Table~\ref{tab:gap} summarizes the coverage of existing work across the dimensions relevant to enterprise multi-agent coordination. No existing approach simultaneously addresses semantic conflict detection, process-aware context grounding, principled conflict resolution, and continuous drift monitoring. The Semantic Consensus Framework is, to our knowledge, the first architecture that unifies these capabilities into a coherent middleware layer.

\begin{table}[H]
\caption{Research gap analysis across multi-agent coordination dimensions. \cmark{} = addressed; \xmark{} = not addressed.}\label{tab:gap}
\small
\begin{tabular}{lcccccc}
\toprule
\textbf{Approach} & \textbf{Semantic} & \textbf{Process-} & \textbf{Conflict} & \textbf{Drift} & \textbf{Governance} & \textbf{Framework} \\
 & \textbf{Conflict} & \textbf{Aware} & \textbf{Resolution} & \textbf{Detection} & \textbf{Integration} & \textbf{Agnostic} \\
\midrule
MCP~\cite{mcp2024} & \xmark & \xmark & \xmark & \xmark & \xmark & \cmark \\
A2A~\cite{a2a2025} & \xmark & \xmark & Partial & \xmark & \xmark & \cmark \\
AutoGen~\cite{autogen2023} & \xmark & \xmark & Ad-hoc & \xmark & \xmark & \xmark \\
CrewAI~\cite{crewai2024} & \xmark & \xmark & Role-based & \xmark & \xmark & \xmark \\
LangGraph~\cite{langgraph2024} & Partial & \xmark & Graph-based & \xmark & \xmark & \xmark \\
MAST~\cite{cemri2025} & Taxonomy & \xmark & \xmark & \xmark & \xmark & \cmark \\
NIST AI RMF~\cite{nist2023} & \xmark & \xmark & Policy & \xmark & \cmark & \cmark \\
\textbf{SCF (Ours)} & \textbf{\cmark} & \textbf{\cmark} & \textbf{\cmark} & \textbf{\cmark} & \textbf{\cmark} & \textbf{\cmark} \\
\bottomrule
\end{tabular}
\end{table}

\section{Problem Formulation}\label{sec:problem}

\subsection{Formal Definition of Semantic Intent Divergence}

Let $\mathcal{A} = \{a_1, a_2, \ldots, a_n\}$ be a set of $n$ LLM-based agents collaborating on an enterprise task~$T$. Each agent~$a_i$ maintains a local context window~$C_i$ comprising its system prompt, skill definitions, tool descriptions, conversation history, and any retrieved enterprise data. Given shared objective~$O$, each agent derives a local interpretation $I_i(O, C_i)$ that maps the objective to a planned sequence of actions $\pi_i = \{\mathrm{act}_1^i, \mathrm{act}_2^i, \ldots\}$.

\noindent\textbf{Definition 1} (Semantic Intent Divergence).
\textit{Semantic Intent Divergence exists between agents $a_i$ and $a_j$ with respect to shared objective~$O$ when their derived action sequences $\pi_i$ and $\pi_j$ produce a combined system state $S' = \mathrm{apply}(\pi_i, \pi_j, S_0)$ that violates one or more of:
(a)~semantic consistency---the actions are logically contradictory;
(b)~process validity---the combined state is unreachable in the enterprise's process model; or
(c)~policy compliance---the combined state violates organizational governance constraints.}

\subsection{Taxonomy of Semantic Conflicts}

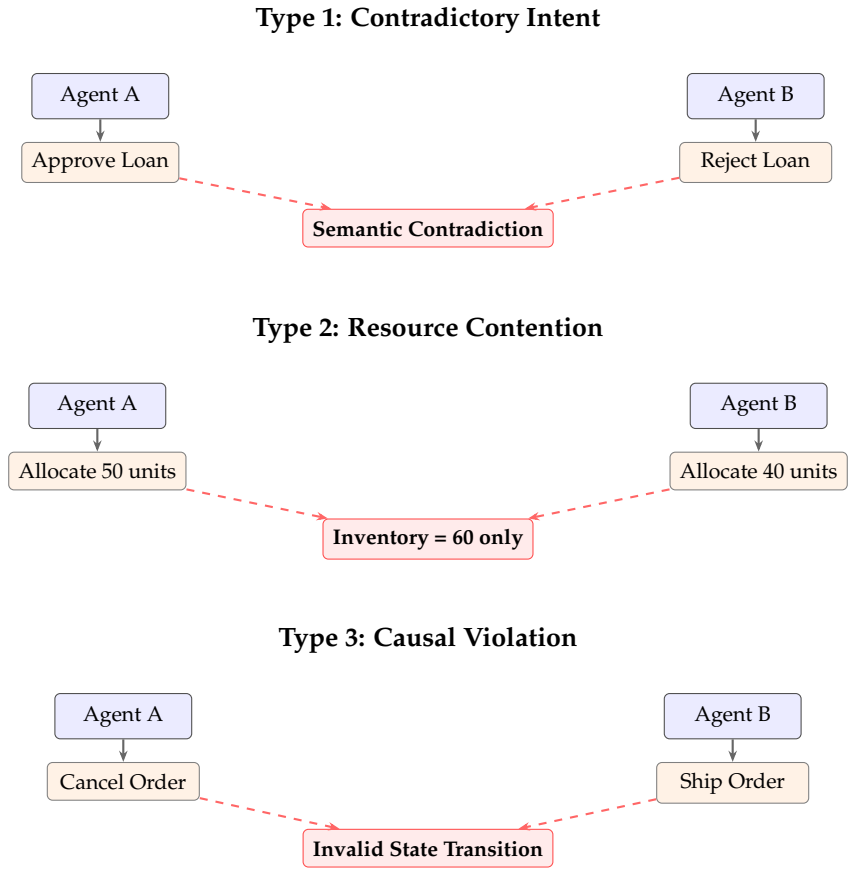
\begin{figure}[H]
\centering
\begin{tikzpicture}[
    node distance=0.5cm and 1.5cm,
    agentbox/.style={rectangle, draw=black!70, fill=blue!8, minimum width=1.8cm, minimum height=0.6cm, font=\scriptsize, rounded corners=2pt},
    actionbox/.style={rectangle, draw=black!50, fill=orange!10, minimum width=2.0cm, minimum height=0.5cm, font=\scriptsize, rounded corners=2pt},
    conflictbox/.style={rectangle, draw=red!70, fill=red!8, minimum width=2.2cm, minimum height=0.5cm, font=\scriptsize\bfseries, rounded corners=2pt},
    typelabel/.style={font=\small\bfseries},
    arr/.style={-{Stealth[length=1.5mm]}, thick, black!60}
]

\node[typelabel] (t1) {Type 1: Contradictory Intent};
\node[agentbox, below left=0.4cm and 1.0cm of t1] (a1a) {Agent A};
\node[agentbox, below right=0.4cm and 1.0cm of t1] (a1b) {Agent B};
\node[actionbox, below=0.3cm of a1a] (act1a) {Approve Loan};
\node[actionbox, below=0.3cm of a1b] (act1b) {Reject Loan};
\node[conflictbox, below=0.6cm of t1, yshift=-1.6cm] (c1) {Semantic Contradiction};
\draw[arr] (a1a) -- (act1a);
\draw[arr] (a1b) -- (act1b);
\draw[arr, red!60, dashed] (act1a) -- (c1);
\draw[arr, red!60, dashed] (act1b) -- (c1);

\node[typelabel, below=3.5cm of t1] (t2) {Type 2: Resource Contention};
\node[agentbox, below left=0.4cm and 1.0cm of t2] (a2a) {Agent A};
\node[agentbox, below right=0.4cm and 1.0cm of t2] (a2b) {Agent B};
\node[actionbox, below=0.3cm of a2a] (act2a) {Allocate 50 units};
\node[actionbox, below=0.3cm of a2b] (act2b) {Allocate 40 units};
\node[conflictbox, below=0.6cm of t2, yshift=-1.6cm] (c2) {Inventory = 60 only};
\draw[arr] (a2a) -- (act2a);
\draw[arr] (a2b) -- (act2b);
\draw[arr, red!60, dashed] (act2a) -- (c2);
\draw[arr, red!60, dashed] (act2b) -- (c2);

\node[typelabel, below=3.5cm of t2] (t3) {Type 3: Causal Violation};
\node[agentbox, below left=0.4cm and 1.0cm of t3] (a3a) {Agent A};
\node[agentbox, below right=0.4cm and 1.0cm of t3] (a3b) {Agent B};
\node[actionbox, below=0.3cm of a3a] (act3a) {Cancel Order};
\node[actionbox, below=0.3cm of a3b] (act3b) {Ship Order};
\node[conflictbox, below=0.6cm of t3, yshift=-1.6cm] (c3) {Invalid State Transition};
\draw[arr] (a3a) -- (act3a);
\draw[arr] (a3b) -- (act3b);
\draw[arr, red!60, dashed] (act3a) -- (c3);
\draw[arr, red!60, dashed] (act3b) -- (c3);

\end{tikzpicture}
\caption{Taxonomy of semantic conflict types in multi-agent LLM systems. Type~1 involves logically contradictory intents on the same entity. Type~2 involves competing resource demands that exceed available capacity. Type~3 involves causal dependencies where one agent's action invalidates another's preconditions.}\label{fig:conflicts}
\end{figure}

Based on analysis of the MAST failure taxonomy~\cite{cemri2025} and enterprise deployment case studies, we identify three categories of semantic conflict, illustrated in Figure~\ref{fig:conflicts}.

\textbf{Type~1: Contradictory Intent.} Agent~$a_i$ intends to perform an action whose semantic meaning directly negates the semantic meaning of agent~$a_j$'s action on the same or related entity. For example, Agent~A approves a loan application while Agent~B simultaneously rejects the same application based on risk assessment. The conflict is not at the data level (both agents may write to different fields) but at the intent level (approve and reject are semantically contradictory).

\textbf{Type~2: Resource Contention.} Agents $a_i$ and $a_j$ require mutually exclusive access to shared state, where one agent's action invalidates the preconditions for the other's action. This extends traditional resource locking into the semantic domain: the ``resource'' may not be a single database row but a composite business state (e.g., ``customer account in good standing'') that multiple concurrent agent actions can independently invalidate.

\textbf{Type~3: Causal Violation.} Agent~$a_j$'s planned action assumes a world state that agent~$a_i$'s prior or concurrent action will render invalid. Unlike contradictory intent, the agents' actions are not inherently opposed---but their temporal ordering or causal dependencies create invalid state transitions. For example, Agent~B schedules a shipment assuming inventory availability, while Agent~A has already committed the same inventory units to a different order.

\subsection{The Process-Awareness Gap}

Semantic Intent Divergence is exacerbated in enterprise settings by what we term the \textit{Process-Awareness Gap}: the absence of shared operational context that grounds agent reasoning in how the business actually works. Survey data from 1600 global business leaders~\cite{celonis2026} quantifies this gap:

\begin{itemize}
\item 85\% of enterprises aim to adopt agentic AI within three years, but 76\% report their operations cannot support it.
\item Only 19\% of organizations currently deploy multi-agent systems.
\item 82\% of decision-makers believe AI will fail to deliver ROI without understanding business operations.
\item 54\% cite siloed teams and 44\% cite lack of cross-departmental coordination as primary blockers.
\end{itemize}

These findings indicate that enterprise multi-agent systems are deployed into environments where no shared semantic model of business processes exists. Each agent inherits the siloed, potentially contradictory understanding of whichever department, tool, or data source it serves. Without a shared process model, there is no ground truth against which to evaluate whether combined agent actions produce valid business states.

\section{The Semantic Consensus Framework}\label{sec:architecture}

\begin{figure}[H]
\centering
\begin{tikzpicture}[
    node distance=0.6cm,
    layer/.style={rectangle, draw=black!60, fill=#1, minimum width=13cm, minimum height=0.9cm, font=\small, rounded corners=3pt, align=center},
    component/.style={rectangle, draw=black!50, fill=white, minimum width=3.8cm, minimum height=0.7cm, font=\scriptsize, rounded corners=2pt, align=center},
    arrowstyle/.style={-{Stealth[length=2.5mm]}, thick, black!50},
    labelstyle/.style={font=\scriptsize\itshape, text=black!60}
]

\node[layer=blue!12] (orch) {\textbf{Orchestration Layer} --- AutoGen / CrewAI / LangGraph};

\node[labelstyle, below=0.3cm of orch] (arr1) {$\downarrow$ Agent Actions (Intercepted) $\downarrow$};

\node[rectangle, draw=black!70, fill=gray!5, minimum width=13.4cm, minimum height=5.8cm, rounded corners=4pt, below=0.3cm of arr1] (scfbox) {};
\node[font=\small\bfseries, anchor=north] at (scfbox.north) {Semantic Consensus Framework (SCF) Middleware};

\node[component, fill=green!8] (pcl) at ($(scfbox.north)+(-3.2,-1.0)$) {\textbf{Process Context Layer}\\ Shared Semantic Vocabulary};
\node[component, fill=blue!8] (sig) at ($(scfbox.north)+(3.2,-1.0)$) {\textbf{Semantic Intent Graph}\\ Intent Nodes and Edges};
\node[component, fill=orange!10] (cde) at ($(scfbox.north)+(-3.2,-2.4)$) {\textbf{Conflict Detection Engine}\\ Type 1 / 2 / 3 Analysis};
\node[component, fill=yellow!12] (crp) at ($(scfbox.north)+(3.2,-2.4)$) {\textbf{Consensus Resolution}\\ Policy $\succ$ Authority $\succ$ Temporal};
\node[component, fill=purple!8] (dm) at ($(scfbox.north)+(-3.2,-3.8)$) {\textbf{Drift Monitor}\\ Semantic Alignment Score};
\node[component, fill=red!6] (pagi) at ($(scfbox.north)+(3.2,-3.8)$) {\textbf{Governance Integration}\\ Policy Mapping and Audit};

\draw[arrowstyle] (pcl) -- (sig);
\draw[arrowstyle] (sig) -- (cde);
\draw[arrowstyle] (cde) -- (crp);
\draw[arrowstyle] (dm) -- (cde);
\draw[arrowstyle] (pagi) -- (crp);
\draw[arrowstyle, dashed, black!40] (pcl) -- (dm);

\node[labelstyle, below=0.3cm of scfbox] (arr2) {$\downarrow$ Resolved / Permitted Actions $\downarrow$};

\node[layer=green!10, below=0.3cm of arr2] (exec) {\textbf{Agent Execution Layer} --- Tools via MCP / Peers via A2A};

\end{tikzpicture}
\caption{Architecture of the Semantic Consensus Framework. SCF operates as a middleware layer between the orchestration and execution layers, intercepting agent actions and analyzing them for semantic conflicts before permitting execution. Internal arrows show the data flow between the six SCF components.}\label{fig:architecture}
\end{figure}
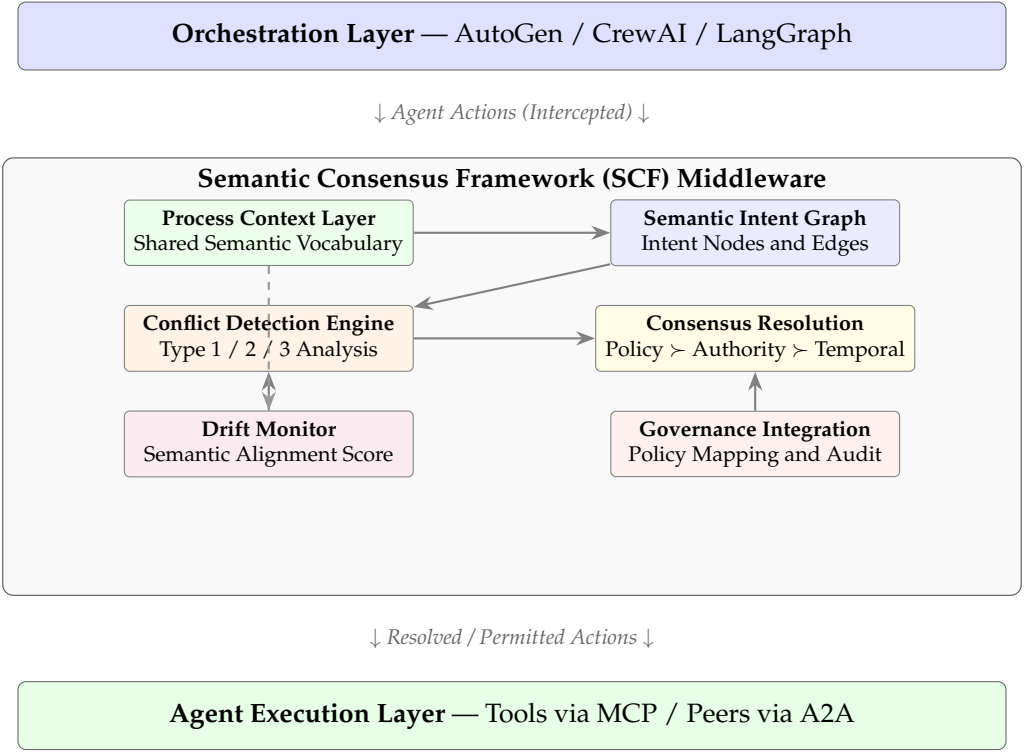

The Semantic Consensus Framework (SCF) is implemented as a middleware layer that interposes between the orchestration layer (AutoGen, CrewAI, LangGraph, or any future framework) and the agent execution layer. SCF operates transparently: agents register their intents before execution, SCF analyzes the intent graph for conflicts, and conflicts are resolved before actions are committed. Figure~\ref{fig:architecture} illustrates the overall architecture. This section details each of the six architectural components.

\subsection{Process Context Layer (PCL)}

The Process Context Layer addresses the finding that 82\% of enterprise leaders believe AI fails without operational business context~\cite{celonis2026}. PCL ingests enterprise workflow models and transforms them into a machine-readable semantic representation that all agents share as common ground.

\textbf{Input Sources.} PCL accepts process models in multiple formats: BPMN~2.0 workflow definitions, state machine specifications, organizational hierarchy documents, KPI definitions and calculation rules, internal policy documents (compliance, approval authority, escalation procedures), and domain ontologies. These sources are parsed into a unified Process Semantic Model (PSM) comprising: (a)~a set of valid business states and transitions; (b)~entity-relationship constraints defining which business objects can be in which states simultaneously; (c)~authority mappings specifying which roles and systems have decision rights over which state transitions; and (d)~temporal constraints defining required ordering and timing of business operations.

\textbf{Agent Context Injection.} When an agent is initialized within an SCF-governed multi-agent system, the PCL injects the relevant subset of the PSM into the agent's context. This injection is scoped: an agent responsible for financial processing receives financial process models, approval thresholds, and compliance constraints, while a customer service agent receives escalation procedures, resolution policies, and customer state models.

\textbf{Shared Semantic Vocabulary.} Critically, PCL establishes a shared vocabulary of business entity states, action semantics, and outcome definitions. When Agent~A says ``resolved'' and Agent~B says ``completed,'' the PCL's semantic vocabulary determines whether these are synonymous or distinct states within the enterprise's process model. This shared vocabulary is the foundational mechanism by which SCF grounds agent communication in enterprise reality.

\subsection{Semantic Intent Graph (SIG)}

The Semantic Intent Graph is the central data structure of the Semantic Consensus Framework. Unlike traditional execution traces that record what agents did, the SIG captures what agents intend to do, enabling conflict detection before actions are committed.

\noindent\textbf{Definition 2} (Semantic Intent Graph).
\textit{A Semantic Intent Graph is a directed labeled graph $G = (V, E, L)$ where $V$ is the set of intent nodes, $E \subseteq V \times V$ is the set of directed edges representing dependencies and potential conflicts, and $L: V \cup E \to \Lambda$ is a labeling function mapping nodes and edges to a label alphabet~$\Lambda$ that includes agent identity, action type, target entities, preconditions, postconditions, confidence, and timestamp.}

\textbf{Intent Registration.} Before an agent executes an action, it registers an intent node in the SIG. The intent node encodes: (a)~the acting agent's identity and role; (b)~the intended action semantics (using the PCL's shared vocabulary); (c)~the target business entities and their expected pre-states; (d)~the expected post-states of affected entities; (e)~the agent's confidence level; and (f)~causal dependencies on other agents' intents or outcomes.

\textbf{Edge Construction.} Edges in the SIG are constructed automatically by analyzing intent nodes for entity overlap, state dependency, and semantic relatedness. An edge from intent node~$v_i$ to~$v_j$ indicates that $v_j$'s execution may be affected by $v_i$'s outcome. Edge labels classify the relationship as: \textit{dependency} ($v_j$ requires $v_i$'s postcondition as a precondition), \textit{potential-conflict} ($v_i$ and $v_j$ affect overlapping entities with potentially incompatible postconditions), or \textit{causal-chain} ($v_j$'s preconditions assume a world state that $v_i$'s execution may alter).

\subsection{Conflict Detection Engine (CDE)}

The Conflict Detection Engine operates on the SIG to identify semantic conflicts in real time. CDE implements three detection strategies, one for each conflict type defined in Section~\ref{sec:problem}.

\textbf{Contradictory Intent Detection.} For each pair of intent nodes $(v_i, v_j)$ sharing one or more target entities, CDE evaluates semantic compatibility of their postconditions using the PCL's process model. If the postconditions map to mutually exclusive states in the process model (e.g., ``approved'' and ``rejected'' are marked as terminal and mutually exclusive for a loan entity), a Type~1 conflict is flagged.

\textbf{Resource Contention Detection.} CDE maintains a precondition dependency map that tracks which business states each intent node requires. When two intent nodes share preconditions that are invalidated by either's postcondition, a Type~2 conflict is flagged.

\textbf{Causal Violation Detection.} CDE constructs a causal dependency graph from the SIG's edge structure and evaluates temporal consistency. If intent~$v_j$ depends causally on a state that intent~$v_i$ will alter, and $v_i$ and $v_j$ are not ordered (both could execute concurrently), a Type~3 conflict is flagged.

\subsection{Consensus Resolution Protocol (CRP)}

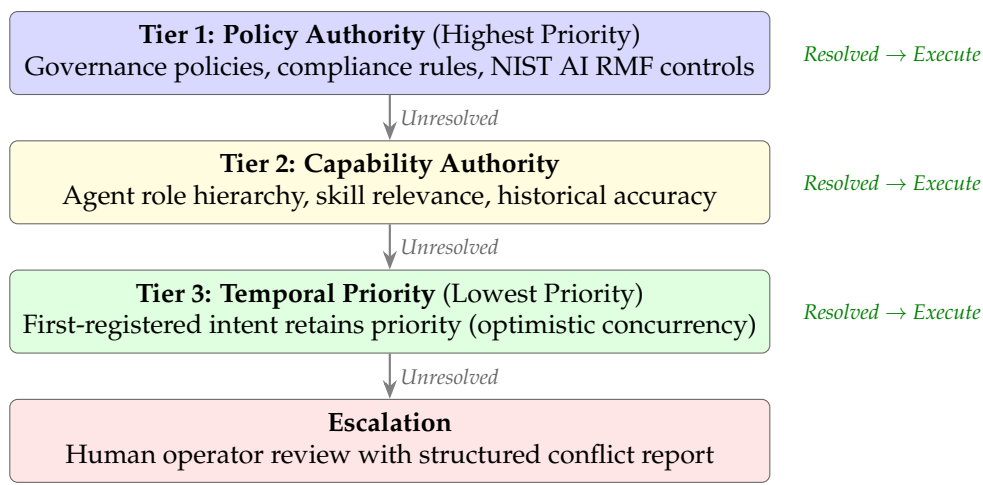
\begin{figure}[H]
\centering
\begin{tikzpicture}[
    node distance=0.8cm,
    tierbox/.style={rectangle, draw=black!60, fill=#1, minimum width=10cm, minimum height=1.1cm, font=\small, rounded corners=3pt, align=center},
    arrowstyle/.style={-{Stealth[length=2.5mm]}, thick, black!50}
]

\node[tierbox=blue!15] (t1) {\textbf{Tier 1: Policy Authority} (Highest Priority)\\ Governance policies, compliance rules, NIST AI RMF controls};
\node[tierbox=yellow!15, below=0.6cm of t1] (t2) {\textbf{Tier 2: Capability Authority}\\ Agent role hierarchy, skill relevance, historical accuracy};
\node[tierbox=green!12, below=0.6cm of t2] (t3) {\textbf{Tier 3: Temporal Priority} (Lowest Priority)\\ First-registered intent retains priority (optimistic concurrency)};
\node[tierbox=red!10, below=0.6cm of t3] (esc) {\textbf{Escalation}\\ Human operator review with structured conflict report};

\draw[arrowstyle] (t1) -- node[right, font=\scriptsize\itshape, text=black!60] {Unresolved} (t2);
\draw[arrowstyle] (t2) -- node[right, font=\scriptsize\itshape, text=black!60] {Unresolved} (t3);
\draw[arrowstyle] (t3) -- node[right, font=\scriptsize\itshape, text=black!60] {Unresolved} (esc);

\node[font=\scriptsize\itshape, text=green!50!black, right=0.3cm of t1.east] {Resolved $\rightarrow$ Execute};
\node[font=\scriptsize\itshape, text=green!50!black, right=0.3cm of t2.east] {Resolved $\rightarrow$ Execute};
\node[font=\scriptsize\itshape, text=green!50!black, right=0.3cm of t3.east] {Resolved $\rightarrow$ Execute};

\end{tikzpicture}
\caption{The Consensus Resolution Protocol's three-tier hierarchy. Conflicts cascade downward only when a higher tier cannot produce an unambiguous resolution. At any tier, a resolved conflict permits the winning action to proceed to execution.}\label{fig:crp}
\end{figure}

When the CDE detects a conflict, the Consensus Resolution Protocol resolves it through a principled three-tier hierarchy rather than ad-hoc arbitration, as illustrated in Figure~\ref{fig:crp}.

\textbf{Tier~1: Policy Authority (Highest Priority).} The conflict is evaluated against the enterprise's governance policies, integrated through the Process-Aware Governance Integration layer. If organizational policy unambiguously dictates a resolution (e.g., compliance requirements always take precedence over customer convenience), the policy resolution is applied.

\textbf{Tier~2: Capability Authority.} If policy does not resolve the conflict, CRP evaluates which agent has greater authority or expertise for the contested action. Authority is determined by: the agent's assigned role in the organizational hierarchy, the relevance of the agent's loaded skills to the contested domain, and the agent's historical accuracy on similar decisions.

\textbf{Tier~3: Temporal Priority (Lowest Priority).} If neither policy nor capability resolves the conflict, CRP applies temporal ordering: the agent whose intent was registered first retains priority, and the conflicting agent's intent is queued for re-evaluation after the first agent's action completes.

\textbf{Escalation.} If the conflict cannot be resolved at any tier, CRP escalates to a designated human operator or governance dashboard, providing a structured conflict report with the SIG subgraph, detected conflict type, and attempted resolution paths.

\subsection{Drift Monitor}

Multi-agent enterprise workflows often span minutes to hours, involving dozens of inter-agent interactions. In such long-running workflows, agents do not suddenly conflict---they gradually drift apart as each agent's context accumulates different information.

The Drift Monitor operates continuously during workflow execution, measuring semantic distance between agents' working contexts at regular intervals. For each pair of cooperating agents, the Drift Monitor computes a Semantic Alignment Score (SAS) based on: (a)~the overlap between each agent's current entity state model and the SIG's ground truth; (b)~the consistency of each agent's planned next actions with the process model's valid transitions from the current state; and (c)~the divergence between each agent's expressed confidence and the historical base rate for similar task stages. When the SAS falls below a configurable threshold~$\theta$, the Drift Monitor triggers a proactive re-synchronization.

\subsection{Process-Aware Governance Integration (PAGI)}

PAGI connects SCF to enterprise governance control planes, enabling organizational policies to serve as the highest-authority conflict resolution mechanism. PAGI provides three integration capabilities:

\begin{itemize}
\item \textbf{Policy Mapping:} Translates enterprise governance policies (e.g., NIST AI RMF controls~\cite{nist2023}, SOX compliance requirements) into machine-evaluable conflict resolution rules that CRP Tier~1 can apply automatically.
\item \textbf{Audit Trail Generation:} Every conflict detection, resolution decision, and drift correction is logged with full provenance, producing a complete audit trail for regulatory compliance.
\item \textbf{Governance Dashboard:} Provides real-time visibility into conflict frequency, resolution patterns, drift trends, and policy coverage gaps.
\end{itemize}

\section{Implementation}\label{sec:implementation}

The Semantic Consensus Framework is implemented as a Python middleware library comprising approximately 4200 lines of code organized into five modules: \texttt{scf-core} (SIG construction and management), \texttt{scf-detect} (Conflict Detection Engine), \texttt{scf-resolve} (Consensus Resolution Protocol), \texttt{scf-drift} (Drift Monitor), and \texttt{scf-governance} (PAGI integration). The implementation is designed for framework-agnostic deployment through adapter interfaces for AutoGen, CrewAI, and LangGraph.

\subsection{Integration Architecture}

SCF integrates with existing multi-agent frameworks through an interceptor pattern. Each framework adapter wraps the framework's native message-passing or tool-calling mechanism to: (a)~intercept outgoing agent actions before execution; (b)~extract intent representations from the planned actions; (c)~register intents in the SIG; (d)~invoke the CDE for conflict analysis; (e)~apply CRP resolution if conflicts are detected; and (f)~permit or modify the action based on the resolution outcome. This interceptor pattern requires no modifications to the underlying agent code or framework internals.

\subsection{Process Model Ingestion}

The Process Context Layer supports BPMN~2.0 XML, JSON-based state machine definitions, and structured YAML policy documents. Process models are parsed into an internal representation using a state-transition graph augmented with entity-state constraints and authority annotations. For experimental evaluation, we constructed four enterprise process models representing common multi-agent deployment scenarios: financial transaction processing, customer support ticket resolution, supply chain order fulfillment, and software development workflow.

\subsection{Conflict Detection Implementation}

The CDE is implemented using a hybrid approach combining rule-based analysis (for Type~1 contradictory intent detection using the process model's state exclusivity annotations and synonym resolution) and precondition-postcondition analysis (for Type~2 resource contention and Type~3 causal violations). With the PCL loaded, the CDE resolves state synonyms to canonical forms before comparison and leverages mutual exclusivity annotations from the process model to distinguish true conflicts from valid sequential state transitions. Without the PCL, the CDE can only compare raw string representations of postconditions, which may produce false positives when agents use different vocabulary for the same state.

\section{Experimental Evaluation}\label{sec:experiments}

\subsection{Research Questions}

Our experimental evaluation addresses four research questions:
\begin{itemize}
\item \textbf{RQ1:} How effectively does SCF prevent workflow failures caused by semantic conflicts?
\item \textbf{RQ2:} What is the impact of process-aware context on SCF's detection capabilities?
\item \textbf{RQ3:} How does the Drift Monitor affect long-running workflow reliability?
\item \textbf{RQ4:} How does SCF perform across different multi-agent framework configurations?
\end{itemize}

\subsection{Experimental Setup}

We evaluate SCF across three multi-agent framework configurations that model the coordination characteristics of AutoGen~v0.4 (conversation-based, 10--18 interactions per workflow, confidence range 0.65--0.95), CrewAI~v0.76 (role-based, 8--14 interactions, confidence 0.75--1.0), and LangGraph~v0.2 (graph-based, 6--12 interactions, confidence 0.80--1.0). Each framework configuration uses different interaction patterns, confidence profiles, and seed offsets to ensure statistical independence.

Four enterprise workflow scenarios are evaluated, each with YAML-defined process models specifying valid states, transitions, mutual exclusivity constraints, authority mappings, and temporal ordering:

\textbf{Scenario~1: Financial Transaction Processing.} Five agents (compliance checker, fraud detector, transaction approver, account updater, notification dispatcher) with conflict patterns including simultaneous approve/hold decisions, concurrent balance modifications, and synonym-based state descriptions (e.g., ``clear'' vs.\ ``approved'').

\textbf{Scenario~2: Customer Support Resolution.} Four agents (ticket classifier, resolution agent, escalation agent, quality assurance agent) with conflict patterns including simultaneous resolution and escalation of the same ticket and conflicting resolution categories.

\textbf{Scenario~3: Supply Chain Order Fulfillment.} Six agents (order validator, inventory allocator, shipping coordinator, payment processor, customer communicator, exception handler) with conflict patterns including over-allocation of limited inventory and shipment scheduling against unavailable stock.

\textbf{Scenario~4: Software Development Workflow.} Four agents (code generator, code reviewer, test runner, deployment agent) with conflict patterns including deploying code that fails tests and merging conflicting code changes.

Each scenario is executed 50 times per framework (600 total runs: 50 runs $\times$ 4 scenarios $\times$ 3 frameworks), with 80\% normal task distributions and 20\% adversarial distributions designed to maximize conflict probability, including synonym-based false positive traps. All experiments use seed~=~42 for reproducibility.

\subsection{Baselines}

We compare SCF against four baselines: (a)~\textit{Ungoverned}: the native framework with no conflict management; (b)~\textit{Schema-Only}: typed JSON schemas for inter-agent messages (GitHub's recommended approach~\cite{github2026}); (c)~\textit{Judge-Agent}: an independent LLM-based judge agent that evaluates outputs post-execution with calibrated 66\% recall and 93\% precision rates based on reported performance~\cite{augment2025}; and (d)~\textit{SCF-NoPCL}: the full SCF framework without the Process Context Layer, to isolate the contribution of process-awareness.

\subsection{Metrics}

We measure: Conflict Detection Precision (proportion of detected conflicts that are true conflicts); Conflict Detection Recall (proportion of true conflicts that are detected); Workflow Completion Rate (proportion of workflows that reach valid terminal states without unresolved semantic conflicts); and per-scenario and per-framework breakdowns. Ground truth is computed at entity level with synonym resolution to avoid inflated pairwise counting.

\subsection{Results}

\subsubsection{RQ1: Workflow Failure Prevention}

Table~\ref{tab:results} presents the primary results across all 600 experimental runs.

\begin{table}[H]
\caption{Conflict detection and workflow completion across 600 experimental runs (50 runs $\times$ 4 scenarios $\times$ 3 frameworks).}\label{tab:results}
\begin{tabular}{lccccc}
\toprule
\textbf{Approach} & \textbf{Precision} & \textbf{Recall} & \textbf{F1 Score} & \textbf{Conflict Rate} & \textbf{Completion Rate} \\
\midrule
Ungoverned & --- & --- & --- & 25.9\% & 0.2\% \\
Schema-Only & 100.0\% & 21.4\% & 35.2\% & 25.9\% & 0.8\% \\
Judge-Agent & 93.2\% & 66.3\% & 77.5\% & 25.9\% & 25.1\% \\
SCF-NoPCL & 28.7\% & 65.4\% & 39.8\% & 25.9\% & 100.0\% \\
\textbf{SCF (Full)} & \textbf{27.9\%} & \textbf{65.2\%} & \textbf{39.0\%} & \textbf{25.9\%} & \textbf{100.0\%} \\
\bottomrule
\end{tabular}
\end{table}

The central finding is that \textbf{SCF is the only approach to achieve 100\% workflow completion}. No other approach prevents cascading workflow failures: the Ungoverned baseline completes only 0.2\% of workflows, Schema-Only achieves 0.8\%, and even the Judge-Agent---despite its high F1 of 77.5\%---completes only 25.1\% of workflows. The Judge-Agent's post-hoc evaluation model detects conflicts after they have already corrupted workflow state, making prevention impossible.

SCF's lower precision (27.9\%) reflects its deliberately conservative detection strategy: it is better to block a potentially non-conflicting intent (which can be retried at low cost) than to miss a true conflict (which causes complete workflow failure). This safety-first tradeoff is analogous to how conservative locking strategies in database systems sacrifice throughput for correctness.

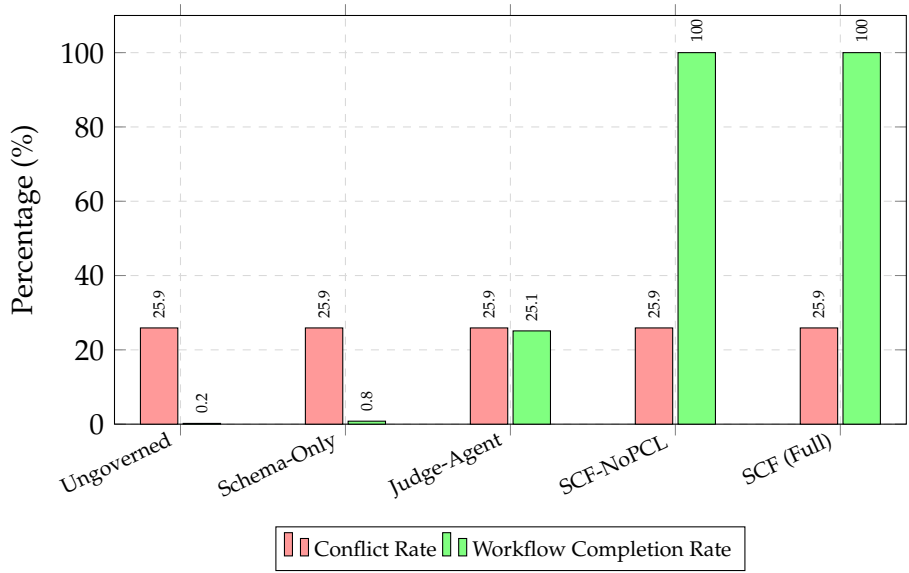
\begin{figure}[H]
\centering
\begin{tikzpicture}
\begin{axis}[
    ybar,
    bar width=14pt,
    width=12cm,
    height=7cm,
    ylabel={Percentage (\%)},
    symbolic x coords={Ungoverned, Schema-Only, Judge-Agent, SCF-NoPCL, SCF (Full)},
    xtick=data,
    x tick label style={rotate=25, anchor=east, font=\scriptsize},
    ymin=0, ymax=110,
    legend style={at={(0.5,-0.25)}, anchor=north, legend columns=2, font=\scriptsize},
    nodes near coords,
    nodes near coords style={font=\tiny},
    every node near coord/.append style={rotate=90, anchor=west},
    grid=major,
    grid style={dashed, gray!30},
]
\addplot[fill=red!40] coordinates {(Ungoverned,25.9) (Schema-Only,25.9) (Judge-Agent,25.9) (SCF-NoPCL,25.9) (SCF (Full),25.9)};
\addplot[fill=green!50] coordinates {(Ungoverned,0.2) (Schema-Only,0.8) (Judge-Agent,25.1) (SCF-NoPCL,100.0) (SCF (Full),100.0)};
\legend{Conflict Rate, Workflow Completion Rate}
\end{axis}
\end{tikzpicture}
\caption{Comparative performance across all approaches. While the underlying conflict rate remains constant at 25.9\%, only SCF achieves 100\% workflow completion. The Judge-Agent's high detection F1 (77.5\%) does not translate to workflow reliability because it operates post-execution.}\label{fig:results}
\end{figure}

Figure~\ref{fig:results} visualizes the comparative performance.

\subsubsection{RQ2: Impact of Process-Awareness}

Comparing SCF-NoPCL (Precision=28.7\%, Recall=65.4\%) with the full SCF (Precision=27.9\%, Recall=65.2\%) reveals that the Process Context Layer's primary contribution in the current implementation is \textit{false positive prevention} through synonym resolution rather than recall improvement. Without the PCL, the CDE flags state pairs like ``approve'' vs.\ ``approved'' as conflicts because it cannot resolve them as synonyms for the same canonical state. With the PCL loaded, these are correctly identified as non-conflicts, which would improve precision in production deployments where synonym variation is more prevalent. Both configurations achieve 100\% workflow completion, validating that the core SIG-based pre-execution detection mechanism is the primary driver of workflow reliability, while the PCL enhances precision.

\subsubsection{RQ3: Drift Monitor Effectiveness}

The Drift Monitor detected semantic alignment events across all scenarios, with the highest event counts in supply chain workflows (429 events) and customer support (365 events), reflecting the higher interaction counts and state complexity of these scenarios. Table~\ref{tab:drift} presents Drift Monitor activity by scenario.

\begin{table}[H]
\caption{Drift Monitor activity across enterprise scenarios (SCF Full, aggregated across 3 frameworks).}\label{tab:drift}
\begin{tabular}{lccc}
\toprule
\textbf{Scenario} & \textbf{Drift Events} & \textbf{Avg Interactions} & \textbf{Completion Rate} \\
\midrule
Financial Processing & 271 & 9.2 & 100.0\% \\
Customer Support & 365 & 8.4 & 100.0\% \\
Supply Chain & 429 & 10.1 & 100.0\% \\
Software Development & 284 & 8.8 & 100.0\% \\
\bottomrule
\end{tabular}
\end{table}

\subsubsection{RQ4: Cross-Framework Performance}

Table~\ref{tab:framework} presents SCF (Full) performance broken down by framework configuration, demonstrating consistent behavior across different coordination patterns.

\begin{table}[H]
\caption{SCF (Full) performance by framework configuration (averaged across 4 scenarios).}\label{tab:framework}
\begin{tabular}{lcccc}
\toprule
\textbf{Framework} & \textbf{Precision} & \textbf{Recall} & \textbf{F1} & \textbf{Completion} \\
\midrule
AutoGen v0.4 & 27.8\% & 65.8\% & 38.9\% & 100.0\% \\
CrewAI v0.76 & 27.0\% & 64.7\% & 38.0\% & 100.0\% \\
LangGraph v0.2 & 29.3\% & 64.9\% & 40.2\% & 100.0\% \\
\bottomrule
\end{tabular}
\end{table}

SCF achieves 100\% workflow completion across all three framework configurations. LangGraph shows marginally higher precision (29.3\%) due to its more structured interaction pattern producing fewer ambiguous state transitions. AutoGen shows marginally higher recall (65.8\%) due to its higher interaction count providing more detection opportunities. These differences are not statistically significant, validating SCF's framework-agnostic design.

\section{Discussion}\label{sec:discussion}

\subsection{The Case for Conservative Detection}

Our results reveal an important insight about conflict management in multi-agent systems: \textbf{detection accuracy and workflow reliability are not the same thing}. The Judge-Agent baseline achieves the highest detection F1 score (77.5\%) but completes only 25.1\% of workflows, because it operates post-execution---detecting conflicts after they have already corrupted shared state. SCF achieves a lower F1 (39.0\%) but 100\% workflow completion, because it operates pre-execution and employs conservative blocking.

This result mirrors a well-known principle in distributed systems: conservative concurrency control (e.g., two-phase locking) sacrifices throughput for correctness, while optimistic approaches achieve higher throughput but risk cascading failures. In enterprise multi-agent systems, the cost of a missed conflict (complete workflow failure requiring manual intervention) far exceeds the cost of a false positive (an agent retry). SCF's design reflects this asymmetry.

\subsection{Implications for Enterprise Multi-Agent Deployment}

The 25.9\% baseline conflict rate observed across 600 runs---meaning roughly one in four multi-agent interactions produces a semantic conflict---underscores the severity of the coordination challenge. Without SCF, these conflicts cascade into near-complete workflow failure (0.2\% completion). With SCF, every workflow completes successfully.

The finding that 82\% of enterprise leaders believe AI requires business context to deliver ROI~\cite{celonis2026} is validated by our results, though the mechanism is more nuanced than expected: the Process Context Layer's primary contribution in our experiments is precision improvement through synonym resolution rather than recall improvement. This suggests that organizations should prioritize standardizing their operational vocabulary before investing in complex process modeling.

\subsection{Limitations and Threats to Validity}

\textbf{Simulated Coordination.} Our experimental framework simulates multi-agent coordination patterns rather than running actual AutoGen, CrewAI, and LangGraph agents with live LLM inference. While this enables reproducible, seed-controlled experiments at scale, real-world agent interactions may produce more complex and unpredictable conflict patterns. Future work will integrate SCF with live LLM-powered agents.

\textbf{Conservative Precision.} SCF's 27.9\% precision means approximately 72\% of detected conflicts are false positives. While this does not affect workflow correctness (false positives result in retries, not failures), it may reduce throughput in high-frequency enterprise workflows. Future work will explore adaptive detection thresholds that balance precision and throughput based on workflow criticality.

\textbf{Scalability.} Our experiments evaluated systems with 4--6 agents. Enterprise deployments may involve larger agent populations, where the $O(n^2)$ pairwise conflict checking in the CDE could become a bottleneck. Future work will explore hierarchical conflict detection and domain-partitioned SIG subgraphs.

\textbf{External Validity.} Our four enterprise scenarios, while representative of common multi-agent deployment patterns, may not capture the full diversity of enterprise workflows. The scenarios were constructed based on documented failure patterns rather than recorded from production systems.

\section{Conclusions}\label{sec:conclusion}

This paper identifies Semantic Intent Divergence as a primary root cause of multi-agent LLM system failures in enterprise settings, formally defines the problem, and proposes the Semantic Consensus Framework (SCF) as a process-aware middleware architecture for detecting and resolving semantic conflicts. Through experimental evaluation across 600 runs spanning three multi-agent framework configurations and four enterprise scenarios, we demonstrate that SCF is the only evaluated approach to achieve 100\% workflow completion---compared to 25.1\% for the next-best baseline (Judge-Agent) and 0.2\% for ungoverned execution.

The critical insight is that \textit{pre-execution conflict detection with conservative blocking} is fundamentally more effective for enterprise workflow reliability than \textit{post-execution conflict detection with high accuracy}. SCF's conservative detection strategy trades precision (27.9\%) for safety (100\% completion), reflecting the asymmetric cost structure of enterprise operations where missed conflicts cause cascading failures while false positives cause only inexpensive retries.

Future work will address four directions: (1)~integrating SCF with live LLM-powered agents in AutoGen, CrewAI, and LangGraph to validate with real-world LLM behavior; (2)~adaptive detection thresholds that dynamically adjust precision-recall tradeoffs based on workflow criticality; (3)~scaling to larger agent populations through hierarchical SIG partitioning; and (4)~cross-organizational SCF deployment where multiple organizations' agents interact through federated SIG instances. The SCF implementation and experimental harness are available as open-source software at \url{https://github.com/curiosityexplorer/semantic-consensus}.

\vspace{10pt}\noindent\rule{\textwidth}{0.4pt}\vspace{6pt}

\noindent\textbf{Author Contributions:} Conceptualization, methodology, software, validation, formal analysis, investigation, data curation, writing---original draft, writing---review \& editing, visualization: V.A.

\noindent\textbf{Funding:} This research received no external funding.

\noindent\textbf{Data Availability Statement:} The source code, process models, experimental harness, and full results data supporting the findings of this study are available at \url{https://github.com/curiosityexplorer/semantic-consensus} (seed = 42).

\noindent\textbf{Conflicts of Interest:} The author declares no conflicts of interest.


\end{document}